# Asynchronous Cellular Operations on Gray Images Extracting Topographic Shape Features and Their Relations

Igor Polkovnikov
Yes San Francisco LLC, ipolk@virtuar.com, www.virtuar.com/ia/
March 19, 2013

## Abstract

A variety of operations of cellular automata on gray images is presented. All operations are of a wave-front nature finishing in a stable state. They are used to extract shape describing gray objects robust to a variety of pattern distortions. Topographic terms are used: "lakes", "dales", "dales of dales". It is shown how mutual object relations like "above" can be presented in terms of gray image analysis and how it can be used for character classification and for gray pattern decomposition. Algorithms can be realized with a parallel asynchronous architecture.

## Keywords

## Introduction

Given an infinite graph **I** where each vertex called an **element** is connected to 8 neighboring elements. Each nine connected elements is a **neighborhood.** Elements in a neighborhood are abbreviated like this, where e5 denotes a **central element**.

| e1 | e2 | e3 |
|----|----|----|
| e4 | e5 or just e | e6 |
| e7 | e8 | e9 |

Elements have a numeric **value**. Additionally, an element may be **marked** in a certain way. An instance of the graph where each element contains a certain value or/and a mark is called an **image**. Marks specify subgraphs. If all elements are marked in the same way or not marked at all, an image is called **gray or single-phased**. If some elements are marked differently - an image is called a **multiphase image**. One of the phases may be called a background. A subgraph may be called an **object**, a subgraph is called **gray** if its elements contain different values.

On an image, a **pattern** is any combination of elements exposing for an observer a certain unity to consider. For example, a human may see three patterns each resembling a letter "A". They do not have to be marked in any way as such or marked at all.

Any element may be put in functional dependency to its own value and values of its neighbors. Let us call these functions e5 = F(e1, e2, e3, e4, e5, e6, e7, e8 ) "**cell-functions**". Note that the element accepting the function value is in the same graph as its neighborhood. This kind of functions is of interest in this article.

Occasionally, we may need a second image **M** which usually contains a copy of I made before some operations on I. Let **em** be a corresponding element of M. Then "cell-functions" are
e5 = F(em, e1, e2, e3, e4, e5, e6, e7, e8 ). em may be called "corresponding_original_element" for clarity.

Let **S-original** be an initial image of the graph. At a certain moment F will be "turned on". If there is a machine modeling the graph and the functions, it would go through the process when each element value may change many times. As it is known in cellular automata, the process may be indefinite. Let us consider functions which result in a finite process. **S-final** is a final image when no element is changing its value. Such process is called **operation** and is determined by a cell-function. Results of operation depend on S-original as well as on M and may vary. There could be no change, a new gray image, zeroing all values or maximizing them, or a multiphase image.

**8 directions** are introduced: "up", "left", "down", "right", "left-up" and so on. Their intuitive meaning is sufficient for now. Also I feel that topographical terminology is very suitable for the description of the gray-level patterns, so there will be "**lakes**" and "**dales**", intuitive meaning of which is clear.

# Chapter 1. Basic operations

**Simple example**

A simple cell-function may look like this:

e = min (e1,e4,e7);
if (e5<e) then
 e5 = e;

An initial S-original and the result S-final of the operation is shown here on the same picture. Both are gray images. Blue, dark area denotes S-original with "I"-pattern. Its values have not changed as the result of this operation. Other values, in yellow, light areas represent the result of this operation. Processing other patterns or during other operations, the original values shown on dark may change too. As simple as it is, this function is valuable since, for example, it helps to distinguish between "I"-patterns and "-"-patterns.

**Operation "fill-dales-opened-down-and-lakes"**

Let us define this operation like

el = min (e1,e2,e3,e4);
er = min (e1,e2,e3,e6);
if (e5<el OR e5<er) then e=max(el,er);

and apply it to a "cross"-pattern below. Note that it is a gray single-phase image. The result is below on the right.

**Operation "fill-dales-opened-down-and-lakes" applied to "A"-pattern**

**Operation "remove-dales"**

After the previous operation which have filled the pattern, another function is applied: e = min (e1,e2,e3,e4,e6,e7,e8,e9);
if (e5>e AND e5>corresponding_original_element) then e5 = corresponding_original_element;

This operation removes dales. What is left, is the pattern with lakes filled. See the left picture below.
The difference between this "lake-filled" pattern and the original one is seen on the next picture to the right..
We can say that this is a pattern of the lake of original "A"-pattern. Note that this is two-phase image. Previous images were gray single-phased. This one can be called "gray object on white background" because we know that 0 here represents a phase with all elements equal 0, i.e. "no difference". It is not another unknown "gray" value.

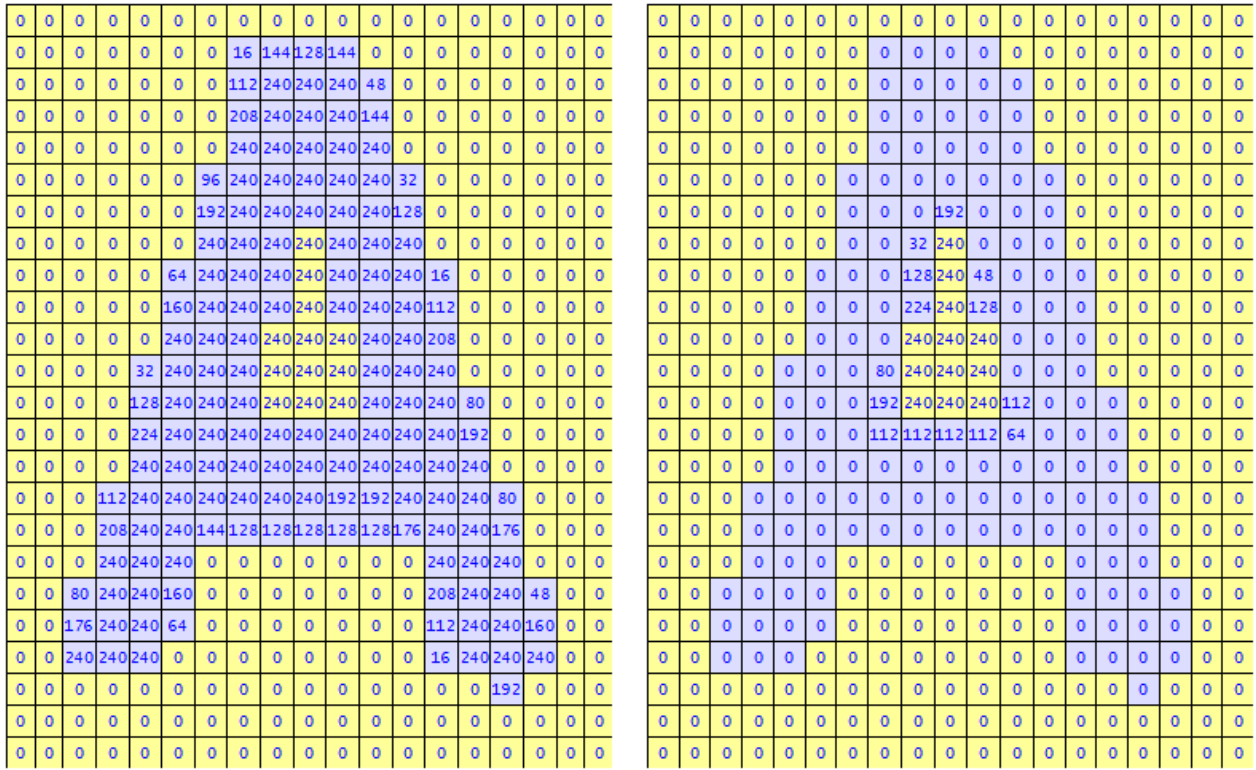

Described operations belong to the class of the "mathematical morphology" operations like erode, dilate, opening, closing, and so on. There are 8 direction on which they can work: d1, d2, d3, d4, d6, d7, d8, d9 ( left, left-top, top, and so on ), named after the neighborhood elements. The reason why I have chosen directional terms over topographical which may sound reasonable since I use "lakes" and "dales" is that usually patterns in human perception are associated with directions: above, below, left, leg, height, horizontal, vertical, and so on.

**Dale as a separate gray object**

Here is the difference of "dale-filled" pattern with the original pattern. If we subtract the "lake" pattern, we obtain the pattern of the dale only. This is going to be a two-phase image "gray-pattern-on-white-background". There is no difficulty to find out that this pattern of the dale is below the lake. "A lake above a dale opened down. This is A"

**Operation with "clean-opened-down" cell function**

As you noticed, the operation "fill-dales-opened-down-and-lakes" fills more than it seems enough for dales to be considered "filled".
Technically, these dales are open not only down, but to some extent left and right. To clean up, to produce more precise dales, dales which are opened only down, not left, right or up, the cell function "clean-opened-down" can be used:

if (e5>corresponding_original_element) then {
  emin = min ( e1, e2, e3, e4, e6 );
 if (emin<e5) then {
  if ( emin > corresponding_original_element ) then
   e5 = emin;
  else
   e5 = corresponding_original_element;
 }
}

This cell function can be explained like this. "If an element is bigger than the corresponding_original_element, then it has been filled as a dale. If surrounding elements neighboring from the direction opposite to the direction of the opening we are interested in are smaller, then the element dale value should be reduced. And it can not be smaller than the corresponding_original_element.". Here is the cross pattern after the operation "fill-dales-opened-down-and-lakes" and consequently after "clean-opened-down".

**Multiple patterns are processed at the same time**

Described operations give result when multiple patterns are present on the same image. Here is an example of such consecutive "fill-dales-opened-down-and-lakes", "clean-opened-down" operations. I want to say that the "lake" and "dale" features of multiple patterns can be extracted at the same time.

**Operation "convex-hull"**

So far in cell functions the operands with 4 elements where used. Let us see what happens with "A"- and "Cross"-patterns when the complete set of 4-element operands is applied. The function looks like this:

d1 = min (e4,e1,e2,e3);
d2 = min (e1,e2,e3,e6);
d3 = min (e2,e3,e6,e9);
d4 = min (e3,e6,e9,e8);
d6 = min (e6,e9,e8,e7);
d7 = min (e9,e8,e7,e4);
d8 = min (e8,e7,e4,e1);
d9 = min (e7,e4,e1,e2);
if (e5<d1 OR e5<d2 OR e5<d3 OR e5<d4 OR e5<d6 OR e5<d7 OR e5<d8 OR e5<d9 ) then
 e5= max (d1,d2,d3,d4,d6,d7,d8,d9);

Essentially, this is a "convex-hull" operation which fills all possible dales turning all patterns into convex hulls. The result is gray. Slight distortion in low pixels ( "a pattern of background") will not change the overall picture: low levels will stay low.

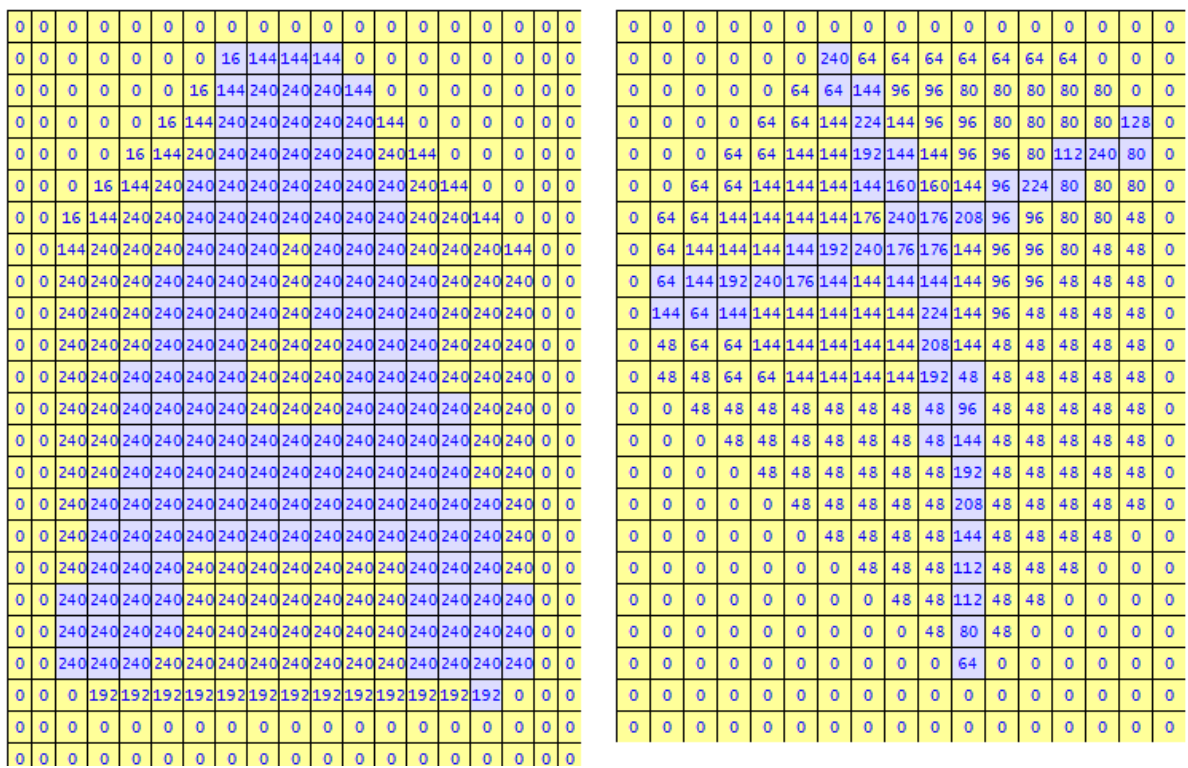

Convex hull operation fills all dales and lakes.

## Chapter 2. Mutual Relations of Objects and Symbol Classification

In the previous chapter, I have shown how to segment out objects which are dales opened in a certain direction and lakes. As the result, since they are separated from the background, we can say that they are marked objects with gray elements. If there are many of them, they can be selected one by one. It is possible to measure histograms of them, areas, perimeters, and so on. These objects are highly indifferent to many distortions of patterns as size, noise, levels, orientation distortions and others. They can be extracted simultaneously at many locations on an image. They are good candidates to be features for pattern recognition. I'd like to introduce a relation which strengthen this claim.

**Mutual Location, Boundary Conditions Stopping the Process**

Once lakes and dales are extracted, is it easy to find their mutual location with element coordinates. But staying in the spirit of this article, I'd like to introduce a kind of "**expansion**" cell-functions which may help to accomplish the same. The function

if (e5<e2) then
 e5 = e2;

will expand down an object or higher values in an image. Expand indefinitely. In order to stop the process, a horizontal line of elements with the highest value should be placed below the area of interest. Above the area of interest, essentially above the object we are interested in to expand, there should be a horizontal line of elements with the lowest value, 0 in our case. The result of the expansion of the previously extracted object "lake" of the "A"-pattern with the mentioned boundary conditions is on the picture. This is "the lake expanded down".

Then, if this image is laid on the image of the "dale-down" object with the operation AND, we have an object which is an intersection. If this object exists, it means that the lake is **above** the dale. Similarly, we can find mutual location in relation to other directions. With simple AND operators on 2 images, it is possible to divide the objects and find not only the existence of relations "above", "below", " to the left", but divide the objects into parts. For example, subtraction of current image from the image of "dale-down" object, will give us 2 objects. Then is it possible to repeat "mutual location" experiments. Also with this approach we may find about subtler relations, for example if a lake wider than the dale. If it is wider, there will be no objects after subtraction. We can find if a lake is "deeper" than a dale and so on..

**Classification of patterns in terms of "lakes", "dales", and their mutual location**

Let us take a quick glance on the latin alphabet letter properties in terms of "lakes", "dales" and their mutual location. The thoughts are summarized in the following table. Some of the more complex features will be considered later. Again, note that all images containing the patterns may be gray and significantly distorted.

| Symbol | Features |
|---|---|
| A | A lake over a dale opened down ( dale-down) |
| B | Two lakes above each other. One small dale-right on the right |
| C | One dale-right |
| D | One lake. A dale-left on the left ( compare with O ). See later |
| E | 2 dales-right |
| F | A dale-right over a dale-down-and-right |
| G | A dale-right with a dale-right. See later |
| H | Dale-up over a dale-down |

| | | |
|---|---|---|
| | I | Dales left-up-down and right-up-down. See the very first cell function |
| | J | Dale-up and -left |
| | K | Tree dales: up, right, down |
| | L | A dale-right and up |
| | M | Three dales: two dales down and one up in between |
| | N | Two dales: one up, another down. They are to the sides of each other ( compare with H ) |
| | O | One lake, no dales |
| | P | One lake above a dale opened right and down |
| | Q | One lake, small dales at the bottom-right |
| | R | One lake, two dales below: one opened right, another down |
| | S | A dale opened right over a dale opened left |
| | T | Two dales to sides of each other, one to right and down, another to left and down. |
| | U | One dale up. |
| | V | One dale up. Two small dales below opened down and to sides ( compare with U)( See O and D ) |
| | W | Two dales up, one down in between. |
| | X | Four dales... |
| | Y | Three dales... |
| | Z | Two dales...directions are contrary to S |

This table should be prolonged to included all other symbols, letters, hieroglyphs.

**Pattern decomposition in relation to lakes and dales.**

Expansion cell-functions can be used to decompose an original pattern into parts. Each part can go through further analysis. For example, below is the way to extract the middle part of the "A"-pattern. The lake object is expanded down and since it is a multiphase image, it is used to binarize the original pattern with a very simple function "if (e-expaned-lake>0) e5=e-original else e5 = 0". On the first picture, an intermediate stage is shown. Since the lake is gray, it may contain remnants of the lake walls. To clean it up, another binarization, now with the not expanded lake object "if (e-lake=0) e5=e-original else e5 = 0"is performed and the result is on the second picture. Note that the resulting images are two-phase. They contain gray objects on the background. The background phase elements here are assigned -1 value instead of 0 for clarity. To finish segmentation of the middle part we have to make similar procedure with the "dale-expanded-down" object.

There should be other ways to achieve the same or similar effect. Similarly, it is possible to remove parts related to lakes and dales from the original image, by, for example, subtraction of images. By expanding a lake (or a dale) to a different direction, we may obtain such things like top, left, "a leg" parts of the original image. This allows more subtle analysis and might be needed to recognize between different styles of the same "A"-pattern.

Expansion can be performed in 1, 2 or 3 directions. For example. On the first picture below, you can see a lake expanded down, up, and right with this cell-function:
if (e5<e2 OR e5<e4 OR e5<e8) then
e5 = max(e2, e4, e8)

This allows extraction of the "left-leg" of the "A"-pattern, by binarization of the first image and subtraction of it from the original image. One may say that this can be achieved without a special cell-function, since it is easy to find coordinates of the lake after it is extracted. Of course, but I want to show how it is done in the framework of this article, without the notion of coordinates. Besides, the operation "expansion" may be performed with the masking from another image in a way similar to what was done in "remove-dales" cell-function. It provides more ways to decompose an image.

Following the method, a logical thing to do is to invert the image. Then, the "leg", which is a "hill" becomes a dale and we can extract it as an object. On the second picture, you can see a part of the original image, inverted, after removing the expanded lake object area. It is a gray image, but removed part is a background "phase". Its elements are assigned to 0 to allow the dale extraction with the same method. The result is on the third image. It contains the gray object "left-leg" on the marked background. Its values are assigned to -1 to highlight that. The extracted dale is a "dale-opened-right", but we could extract a "dale-opened-up" or a "dale-opened-up-and-right'.

**Dales of Dales**

Since the extracted dale is a gray object, we can apply the same methods to it and obtain "dales of dales". Or "lakes of lakes". This might be useful in distinction of such patterns as "C" and "G". "G" pattern has a dale-opened-right of a dale-opened-right, while "C" pattern does not have this feature. The next three pictures show an original "G"-pattern, an object "dale-opened-right" of the original pattern, and an object "dale-opened-right" of the "dale-opened-right". An image on the second picture is a multiphase image, but we ignore it. Since we used 0-value for the background phase, it is possible to apply the same sequence of operations to it.

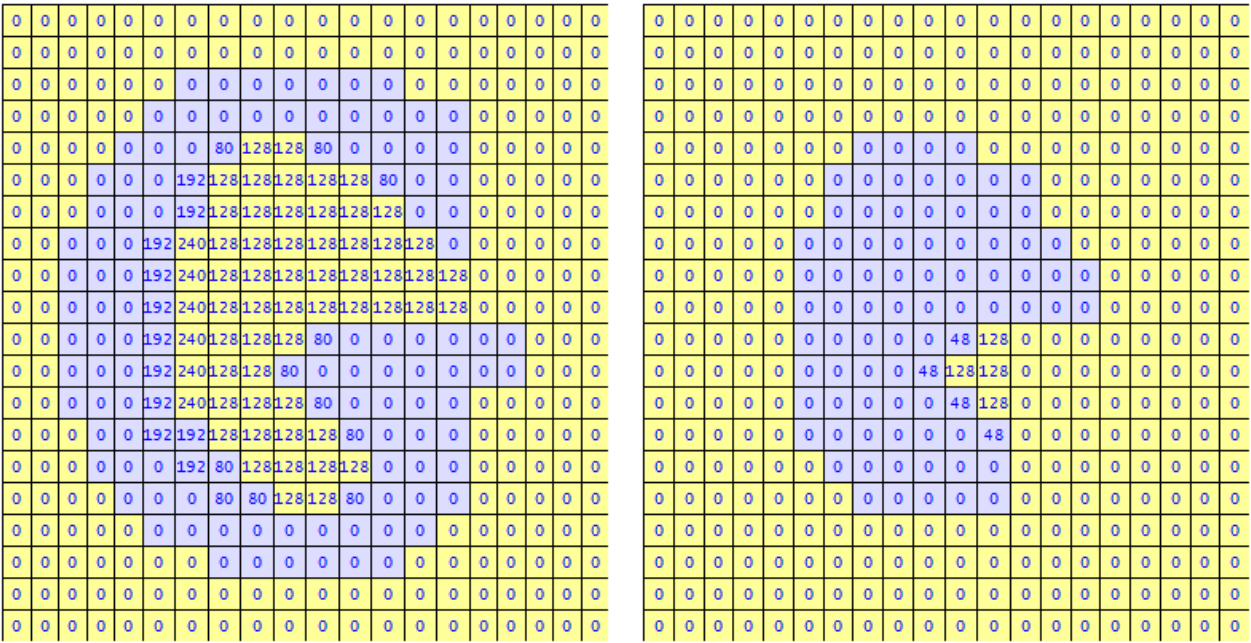

# Chapter 3. Waterfall: Background -> "Objects" -> Lakes -> Islands

For the sake of completeness, I'd like to present the results of a kind of the flooding operations which I will call "waterfall" here:

e = max (e2, e4, e6, e8)
if ( e5 == -1 AND corresponding_original_element <= e) then
e5 = corresponding_original_element

Operation with this cell-function works like this. An image it operates on should be filled with marked elements. The mark used here is -1. Value of these elements is irrelevant. We can say that the image is empty. Marked elements are the elements allowed to change during the operations. Once they are changed, they are unmarked. This is a method allowing dynamically specifying the area of operation. In order to start, some of the elements should have some value, higher than at least one value in the graph which keeps "corresponding_original_elements". In the current example the value 256 is used which is higher than any value in the domain of the original graph which is integers from 0 to 255. Also, in the current example, a value assigned is a value of the corresponding original element, but it can be any value, including 256. This will simplify operations which might follow.

On the first picture below there is an original image of the "A"-pattern which is made a little more interesting. On the second image there is a result of the first run of the operation. Initial conditions of the first graph are the frame of the 256-value elements with inside elements marked -1. All elements outside of the frame, if there are, are not marked. Another graph contains original image. As the result of the operation, all "flooded" elements contain values from the original image, frame stays the same ( it is not visible on the picture ) and elements of the "A"-pattern itself stay marked -1 since they did not change. The area of "-1-marked" elements specify an object area in the original image. This area includes "lakes" if there are. One might say that the valued-elements specify a background of the original image in relation to a certain frame. What is left is an object. Experiment shows that this object is the same as the result of "fill-dales", "remove-dales" consecutive operations.

The third picture depicts the result of the next operations: all valued pixels are assigned to 256, original image is inverted, and the operation waterfall is repeated. So, on the picture, 256 denotes the boundary, "frame" elements, flooded area is filled with [0,255] values ( it happens to be what we would call an "A"-pattern ), and -1 denoted the rest of the elements which must include the lake. The process is repeated again, and the forth picture depicts the result: all elements are valued, none is marked -1. Repeating will not give any more results: there is no "island" on the lake.

# Chapter 4. Spiral pattern: Meanders

Not only analysis of dales of dales is possible, but it is possible to distinguish where a dale "is bent", its meanders speaking geographically. It is presented on the example of the spiral pattern shown on the first picture below. The second picture contains the "convex hull". It is convenient, since it fills all the dales. Then the results starting from

"clean-opened-down" dales, "clean-opened-right", "clean-opened-up" and so on in the contra-clockwise direction are present. All dales can be marked as objects. It is easy to build a tree of the features of a specific dale: an initial "dale-opened-down" contains a "dale-opened-right" contains a "dale-opened-up" contains a "dale-opened-left" and so on. The last picture shows the difference between an image with dales-down filled and cleaned and the original. It is something which might be called an "dale itself". It can be analyzed in the same way. During this, what is analyzed as a dale is an "object itself". Now what is going to be filled and cleaned with the same cell-functions is not a dale, but in fact a "mountain range".

## Addendum

**On application of cell-functions to distinguish between "D" and "O" patterns**

I'd like to present another example showing the difference between "D" and "O" patterns. On the first image there are original patterns. Second image is the result of the operation with

e = min (e3,e6,e9)
if (e5<e) then
e5 = e

which expands patterns left. The third image is the result of the operation with

e = min (e6,e8,e9)
if (e5<e) then
e5 = e;

It expands the pattern to the top-left direction filling the "corners". These two operations on "D" and "O" patterns produce opposite results which may be used in the recognition of these patterns.

# Discussion

The notion of concavity "openness" and its use to describe shapes is known. For example, in [1] the "dales" are called "convexities". Then different convex shapes are classified and used to describe characters. Then shape-based graphs of the character topographic features are introduced and used in recognition of Bengali and Hindi characters. It is stated that this method helps to discriminate characters very similar in shape. Interesting, that a shape denoted with a horizontal line is mentioned often. It corresponds to the shape created by the first cell-function introduced here and used to distinguish between "O" and "D" patterns. The binary images where used though. With the approach proposed in the current article dealing with gray images only, the binarization stage which is very prone to distortions can be omitted.

In [2] An architecture for asynchronous cellular processor is proposed. So called "processor-per-pixel" which is specifically suitable for "wave-front" algorithms. It contains a latch matrix and a combinatorial block per pixel. This is a kind of architecture which can realize operations described in this article very efficiently.

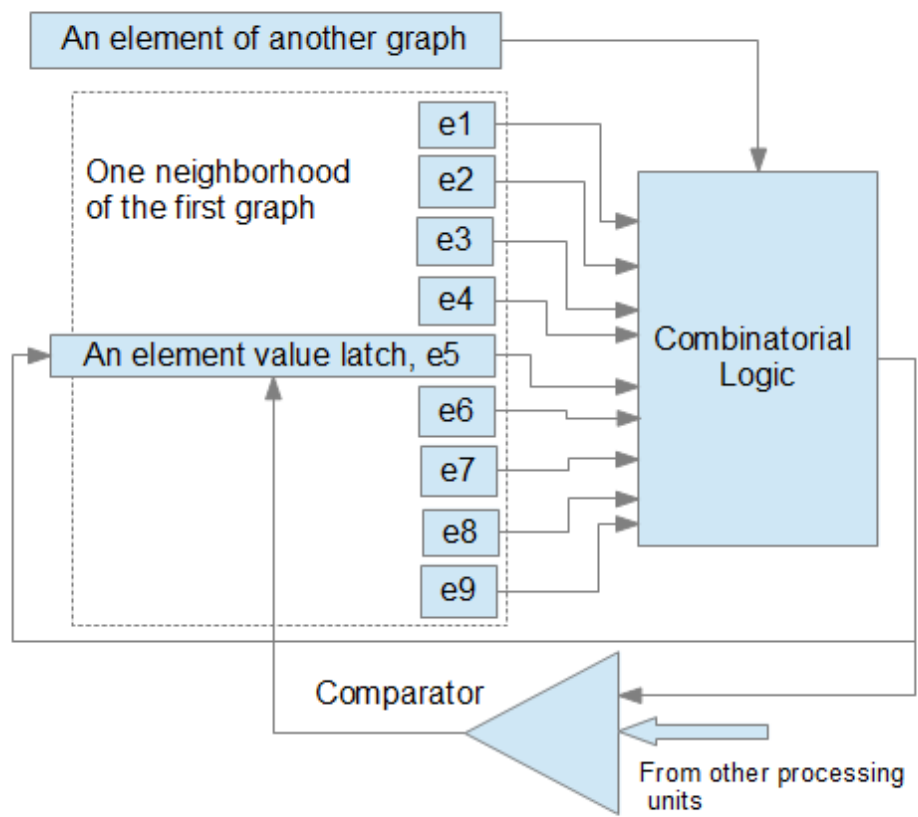

Since the considered operations do not require temporary values to be stored, a block diagram of a processing unit might be like shown on the picture. Combinatorial logic processes values from an element, its neighbors and a corresponding elements of another graph (image matrix) in accordance to a pre-selected cell-function (control logic is not shown). An element value latch during an operation is completely transparent. The end of an operation is defined by a comparator which tracks transient process on all processing units and issues a signal to latch the values in elements when the transient process ends. This is the fastest possible architecture.

If parallel architectures are not considered, it is possible to improve throughput of the algorithm implementation hardware if one complete neighborhood can be obtained via one memory access. The memory design is described here [3].